\documentclass[acmsmall]{acmart}

\AtBeginDocument{%
  \providecommand\BibTeX{{%
    \normalfont B\kern-0.5em{\scshape i\kern-0.25em b}\kern-0.8em\TeX}}}

\usepackage{subfig}
\usepackage[flushleft]{threeparttable}
\usepackage{color, colortbl}
\usepackage{multirow}
\usepackage{array}
\definecolor{Gray}{gray}{0.9}
\usepackage{soul}

\begin{document}

\title[Development and Evaluation of Three Chatbots for Postpartum Mood and Anxiety Disorders]{Development and Evaluation of Three Chatbots for Postpartum Mood and Anxiety Disorders}

\author{Xuewen Yao}
\email{xuewen@utexas.edu}
\affiliation{%
  \institution{The University of Texas at Austin}
  \city{Austin}
  \state{Texas}
  \country{USA}
}

\author{Miriam Mikhelson}
\affiliation{%
  \institution{The University of Texas at Austin}
  \city{Austin}
  \state{Texas}
  \country{USA}
}
\email{mmikhelson@utexas.edu}

\author{S. Craig Watkins}
\email{craig.watkins@austin.utexas.edu}
\affiliation{%
  \institution{The University of Texas at Austin}
  \city{Austin}
  \state{Texas}
  \country{USA}
}

\author{Eunsol Choi}
\email{eunsol@utexas.edu}
\affiliation{%
  \institution{The University of Texas at Austin}
  \city{Austin}
  \state{Texas}
  \country{USA}
}

\author{Edison Thomaz}
\email{ethomaz@utexas.edu}
\affiliation{%
  \institution{The University of Texas at Austin}
  \city{Austin}
  \state{Texas}
  \country{USA}
}

\author{Kaya de Barbaro}
\email{kaya@austin.utexas.edu}
\affiliation{%
  \institution{The University of Texas at Austin}
  \city{Austin}
  \state{Texas}
  \country{USA}
}

\renewcommand{\shortauthors}{Xuewen Yao et al.}

\begin{abstract}
  In collaboration with Postpartum Support International (PSI), a non-profit organization dedicated to supporting caregivers with postpartum mood and anxiety disorders, we developed three chatbots to provide context-specific empathetic support to  postpartum caregivers, leveraging both rule-based and generative models. We present and evaluate the performance of our chatbots using both machine-based metrics and human-based questionnaires. Overall, our rule-based model achieves the best performance, with outputs that are close to ground truth reference and contain the highest levels of empathy. Human users prefer the rule-based chatbot over the generative chatbot for its context-specific and human-like replies. Our generative chatbot also produced empathetic responses and was described by human users as engaging. However, limitations in the training dataset often result in confusing or nonsensical responses. We conclude by discussing practical benefits of rule-based vs. generative models for supporting individuals with mental health challenges. In light of the recent surge of ChatGPT and BARD, we also discuss the possibilities and pitfalls of large language models for digital mental healthcare.

    
\end{abstract}




\maketitle

\section{Introduction}
Postpartum mood and anxiety disorders (PPMADs) affect up to one in five women globally \cite{post_partum_dep} and are associated with many adverse effects on child development across cognitive, motor, and mental health domains \citep{murray_et_al_1996), cumming_etal_1994, goodman_gotlib_1999, gotlib_lee_1996, weissman_etal_2006, Avan690}. Given the ubiquity of PPMADs and their long-term implications for both mothers and children, it is critical to provide treatment as well as support to postpartum women in this period.

Standard treatment for postpartum depression calls for antidepressant medication and psychotherapy \cite{alternative_therapy_ppd}, both of which involve healthcare professionals or trained therapists. Such treatments are not available to all women suffering from PPMADs. This is due to major barriers such as cost, time, stigma, and lack of childcare and information that would allow women to access such mainstream treatments \cite{ppd_barriers}. Nearly 60\% of mothers with PPMAD symptoms are not professionally diagnosed or treated, and 50\% of diagnosed mothers are left untreated, with poor women and women of color being disproportionately impacted \cite{howell_etal_2005}. Additionally, there is a massive global shortage of mental health professionals with approximately 9 psychiatrists per 100,000 people in developed countries and 0.1 for every 1,000,000 in lower-income countries \cite{chatbot_mental_health}.

Alternatives to professional support can also be used to address postpartum mental health concerns. As one example, peer support-- that is, support provided by individuals with lived experience of mental illness -- is gaining recognition as an effective form of mental health support \cite{peer_support_literature_review2}. However, many of the factors limiting the accessibility of professional support to the postpartum population still carry over to peer support, such as time and lack of childcare. Furthermore, any form of support from a provider or peer is dependent on their window of availability overlapping with the support-seeker's time of need. This overlap of availability and need can be challenging to attain, in particular for postpartum parents who may experience distress while awake at night tending to their infants \cite{warmline_night}. Therefore, the aim of this work is to develop an accessible, low-cost, and automated support tool for caregivers at risk of postpartum mood and anxiety disorders that can remedy some of the critical pitfalls of existing forms of support.

Conversational agents, or chatbots, have increasingly become of interest in the healthcare domain for their scalability, wide accessibility, ease of use, and fast information dissemination \citep{covid_bot_review, review_bot_apps, brazil_viki_bot}. Preliminary evidence suggests that chatbots can be successfully used in the domain of mental health to provide psychotherapy  \citep{ survey_psychotherapy_chatbots, cbt_chatbot}, support lifestyle and behavior modification, and psychoeducation \cite{chatbot_mental_health}. In some cases, chatbots may be preferred to human therapists, especially among those who are uncomfortable opening up their feelings with others, such as veterans \citep{chatbot_anonymity, chatbot_mental_health}. Thus, chatbots represent a promising low-cost and low-burden strategy for supporting people with mental health challenges. 

In the current work, we work collaboratively with the non-profit Postpartum Support International (PSI) \cite{psi}, to design, implement, and evaluate three chatbot models with the goal of providing low-cost and accessible support to postpartum support seekers. PSI is the largest international non-profit organization serving the needs of pregnant, post-loss, and postpartum support seekers \cite{psi_largest}. Among various services provided to postpartum support seekers, PSI operates a text-message based ``helpline'' platform where support seekers can chat with trained PSI volunteers. As part of our collaboration with PSI, we were provided access to a dataset of 7014 helpline text-message conversations, which we leveraged to build our chatbots. We undertook several strategies to protect the privacy and anonymity of the support seekers whose conversations we used to develop the chatbots, which we detail in Section~\ref{section: dataset}. 

\subsection{Popular Chatbot Architectures}
From a computational perspective there are two main approaches to building a chatbot: rule-based (retrieval-based) models and generative models. Originally, most chatbots were built with rule-based architecture, such as ELIZA in 1966 \cite{eliza} and PARRY in 1972 \cite{parry}. Rule-based models choose from hard-coded pre-written responses based on a fixed predefined set of rules \cite{overview_chatbot} making it unlikely for them to output irrelevant or inappropriate responses. They are suitable for closed-domain conversations \cite{chatbots_history} and enable quick deployment of an approved set of conversations for information gathering and dissemination \cite{covid_bot_review}. However, as detailed in Section~\ref{sec: c_relatedwork}, a chatbot developed using rule-based models cannot answer questions outside of its preset scope and may generate irrelevant or repeated responses, which can prevent users from engaging in meaningful open-ended conversations. By contrast, generative chatbots appeared in the last decade and have received much critical attention in early 2023 due to the release of large language models such as ChatGPT \cite{chatgpt}. They are typically built systematically using deep learning approaches and a large training dataset \cite{overview_chatbot}. Two architectures, Seq2Seq and Transformer, are commonly used, both of which have an encoder-decoder architecture. However, Transformer-based models have better performance over Seq2Seq-based models due to their self-attention mechanisms \cite{attention}. Generative chatbots allow for a wide range of topics and free-flowing input from users with an overall focus on open-domain dialogue \cite{meena, xiaoice, blender, blenderbot2_1, blenderbot2_2, lamda}. However, as detailed further below, generative models are prone to errors and can output incoherent or irrelevant responses \cite{survey_psychotherapy_chatbots}. Additionally, generative models cannot ``think'': their responses are based on statistics from the training dataset \cite{chatgpt_pinker} and they lack functional linguistic competence \cite{llm_cognitive}, such as formal reasoning, world knowledge, and social reasoning, resulting in errors such as hallucination \cite{hallucination}. This is problematic in the digital mental health domain where targeted users are experiencing emotional distress which can be further exacerbated by insensitive responses to the content they share. However, given recent advances in generative technology \cite{chatgpt, bard, gpt4}, there is overwhelming interest in their potential for healthcare applications including chatbots, and it is reasonable to consider that increasing sophistication of generative models may begin to outweigh their challenges. In this work, we build three chatbots leveraging both rule-based and generative components and evaluate their performance against one another.

\subsection{Chatbot design goals}

The design goals for our chatbots were developed jointly with Postpartum Support International (PSI), who has an existing text message ``helpline'' where support seekers can receive support from volunteers and access to a vast network of resources including local support groups. The overall goal for the proposed chatbot was to supplement these services by providing support-seekers 24/7 access to a naturalistic conversational agent where they could share their current concerns and receive support for them in real time. Specifically, the chatbot would ideally provide i)  empathetic and ii) content-specific responses that validate and address the individual concerns shared by support seekers, as well as iii) actively encourage support seekers to express and share their current challenges. Critically, due to ethical concerns, the chatbot should also iv) refrain from offering advice or resources to support seekers, and v) refer support seekers suffering from severe symptoms, such as suicidal thoughts, to emergency services staffed by humans. We motivate and elaborate these design goals below. 


Our main goal was to design a chatbot that responded empathetically to support seekers' concerns. Empathy refers to the understanding and acknowledgement of what another person is experiencing or trying to express and is fundamental to the success of psychotherapy \citep{empathy_psy,empathy_psy2, empathy}. Previous research has shown that empathetic chatbots help reduce negative mood and comfort support seekers in combating adverse effects of social exclusion \cite{chatbot_empathetic_social_exclusion}. Additionally, chatbots with empathetic behaviors are associated with higher user acceptance and longer interaction times \cite{chatbot_mental_health}. For these reasons, designing an empathetic chatbot was our primary goal. We focused on two specific components of empathy, namely emotional validation and encouragement. Emotional validation reassures the normality of the experience and feelings from a support seeker, such as \textit{I can see how that would be overwhelming}. Encouragement, on the other hand, provides positive and hopeful statements about the future, such as \textit{Hang in there! Things will get easier.}

Content-specific replies are also known to promote user satisfaction. For example, in a pilot study involving 101 participants, content-specific management responses to user-generated hotel reviews gained more trust and delivered higher perceived communication quality than generic ones \cite{ceb_generic_specific}. Similarly, in another study, 176 users rated responses to concerns more favorably when the response included a paraphrase of the complaint, thus making the response more personal and less generic \cite{hotel_generic_specific}. Hence, providing context-specific responses to support seekers should increase their satisfaction while interacting with the chatbots. Finally, Olderbak et al. \cite{emotional_specific_empathy} found that empathy was emotion-specific and measurable, suggesting the possibilities of providing emotion-specific empathy to support seekers reaching out for help.

Additionally, we draw from previous research demonstrating that therapeutic writing and sharing promotes health and healing \citep{writing_pennebaker, writing_healing} and expressive writing can be a helpful early and low-cost universal intervention to prevent postpartum distress for women \cite{expressive_writing_postpartum}. We therefore incorporate open-ended conversation prompts, i.e. open-ended questions, into our bot to provide users the opportunity to engage in expressive writing. 


Next, our designed chatbots should refrain from providing support seekers with advice or resources, and should refer support seekers with severe symptoms to emergency services staffed by humans. Chatbots cannot verify the accuracy of the information they provide or guarantee the helpfulness of their advice \cite{turingadvice}, meaning that providing advice could lead support seekers to feel frustrated. 
Additionally, while a main goal of the PSI helpline staff is to provide access to local resources, we determined with PSI that this would be out of the scope of the current chatbot aims and that resources would be provided via different channels on the platform. 

Finally, and perhaps most importantly, our chatbot needed to be able to identify support seekers suffering from severe symptoms. Prior work with the PSI dataset \cite{descriptive_analysis} has identified that 8.6\% of support seekers express severe symptoms, which comprises suicidal thoughts, self harm, or other indications of safety concerns. Such support seekers may present immediate danger to themselves or others, and chatbots cannot and should not provide the necessary intervention in those situations. Thus, identification of severe symptoms expressed by support seekers, and referral to immediate help from emergency services or trained professionals was determined with PSI to be the optimal course of action for the chatbot.




\subsection{Contributions} 
In this work, we explored both rule-based and generative models to build chatbots for postpartum women. In particular, we seek to answer whether generative models can output high-quality responses given a small training dataset and a vulnerable target population. We also examine whether rule-based models perform better than generative models in this application. The contributions of this work are as follows:

\begin{itemize}
    \item We developed a model to detect the expression of severe symptoms within a sentence from support seekers leveraging sentence embeddings \cite{sbert} and LIWC \cite{liwc} features. Our model successfully detects severe symptoms with an F1 score 0.88 (Precision: 0.91, Recall: 0.85), allowing it to enforce our ethical guidelines for the scope of support the chatbot can provide, i.e., users expressing severe symptoms are redirected to better-suited care provision.
    \item We incorporated empathy, context-specific responses, and open-ended questions to build 3 chatbots that accept open-ended user input to support postpartum women: a baseline model (simple rule-based), a rule-based model (with NLP components), and a generative model. 
    \item We present and evaluate the performance of our chatbots using both machine-based metrics and human evaluations. We found that our rule-based model achieved the best performance overall due to its content-specific and human-like replies.
    \item We conclude with a discussion of our findings that rule-based models (vs. generative models) may be a better choice for supporting people with mental health challenges. In light of the recent surge of work on generative models, e.g. ChatGPT and BARD, we also discuss the possibilities and pitfalls of large language models for digital mental healthcare.
\end{itemize}

\section{Related Work} \label{sec: c_relatedwork}

\subsection{Psychotherapy chatbots}
Psychotherapy chatbots are a class of chatbots designed to broadly address and alleviate mental health challenges. Below, we introduce the reader to several existing psychotherapy chatbots, describing their main features and architecture. 

Over 90\% of existing psychotherapy chatbots rely on rule-based models for their implementation \cite{overview_chatbots_mental_health, survey_psychotherapy_chatbots}. Woebot \cite{woebot} and Wysa \cite{wysa} are two large-scale highly-funded mobile-based psychotherapy chatbots leveraging rule-based techniques. Wysa has served over 400 million conversations to 4.5 million users in 65 countries \cite{wysa_users}, while Woebot exchanges millions of messages with users every week, upwards of 500 million messages to date \cite{woebot_users}. A key strength of these bots is that their design incorporates evidence-based psychotherapy techniques (e.g. Cognitive Behavioral Therapy), to support users with a wide range of mental health needs. Additionally, their effectiveness has been tested using randomized controlled trials that assess changes in mental health before vs. after a period in which users interact with the chatbots \citep{woebot, wysa}. However, they also have some shortcomings as rule-based models. Wysa is unable to answer questions outside of its preset scope and repeated responses are often generated in open topics \cite{survey_psychotherapy_chatbots}. Woebot, on the other hand, is more like a guided interactive Q\&A and may generate irrelevant responses \cite{survey_psychotherapy_chatbots}.

Neither Woebot nor Wysa were specifically designed with the needs of caregivers with PPMADs in mind. However, their broad mental health platforms may be of service to this community. In 2020, Woebot tested the satisfaction and acceptability of the use of Woebot among postpartum women and found that 80\% of the users felt comfortable to use Woebot for mood management \cite{woebot_ppd2}. In January 2023, Woebot began recruiting women who had given birth in the last three months and were experiencing mild to moderate postpartum depression to determine if Woebot can reduce symptoms of depression compared to an educational control in 8 weeks \citep{woebot_ppd, woebot_ppd3}. This initiative brings new opportunities and potential for postpartum women in distress and highlights the growing recognition of this populations' critical need for more accessible support. 

While less common, there are some existing efforts to develop psychotherapy chatbots using generative modeling. This includes Evebot \cite{evebot} and the work by Das et al. \cite{psychotherapy_generative}. Evebot \cite{evebot} is a seq2seq-based chatbot that aims to reduce negative emotions or depression in adolescents. It uses a Bi-LSTM model to detect the negative emotions of users and extract counselling related outputs from an online corpus and a seq2seq model with maximum mutual information (MMI) criterion to ensure the dependence between input and output. Evebot was not designed for multi-turn conversations and has poor performance in both grammar and semantics as limited by the scope of the training dataset. Das et al. \cite{psychotherapy_generative} proposed a novel technique to adapt existing open-domain pre-trained generative models, DialoGPT \cite{dialogpt} and GPT-2 \cite{gpt2}, for therapeutic conversation modeling. Models were fine-tuned using data from subreddit threads on mental health and transcripts of videos on psychotherapy and counseling. However, their model responds with unhealthy advice or generic information usually discouraged in traditional therapeutic counseling, signaling the importance of a high-quality training dataset. 

\subsection{Open-domain chatbots}
A major challenge in building generative chatbots lies in obtaining sufficient, comprehensive, and high-quality training data. Due to the sensitive and private nature of mental health, it is hard to acquire a large psychological or therapeutic corpora that can be used to train successful generative models  \cite{survey_psychotherapy_chatbots}. Conversational agents designed for general purpose input and output, known as open-domain chatbots, do not have this constraint and tend to have better performance. Through substantial efforts and vast amounts of diverse training data, open-domain chatbots have the capacity to engage in open-ended conversations across multiple topics. Below, using examples we illustrate some of the strengths and weaknesses of existing open-domain chatbots leveraging generative architectures. 

Meena \cite{meena} is a social chatbot that is trained on public-domain social-media conversations that can conduct human-like multi-turn conversations. The main architecture is based on a seq2seq model with an evolved transformer. BlenderBot 2.0 \cite{blenderbot2_1, blenderbot2_2} is another chatbot with its own long-term memory and internet access. It can conduct long conversations over multiple sessions as well as search the internet by generating its own search queries, read the results, and take them into account when formulating a response. Additionally, LaMDA (Language Models for Dialog Applications) \cite{lamda} is a transformer-based model developed by Google focused on safety and factual grounding. The LaMDA model ensures that its responses are consistent with a set of human values and provides ways to consult external knowledge sources, such as an information retrieval system, a language translator, and a calculator. These works have made tremendous technical contributions to the field of chatbots, indicating the possibilities and potential for open-ended conversations with generative model architecture. However, these models focused on presenting factual information to a broad set of users as their target population. This is possible through huge amounts of training data and access to the internet but may not be feasible for smaller-scale efforts aiming to develop tools for specific and sensitive populations, such as postpartum mothers. 

\subsubsection{Empathetic Open-domain chatbots}
More adjacent to psychotherapy chatbots, some open-domain generative chatbots have been designed to provide empathetic responses to users. A few notable examples include EMMA \cite{emma}, XiaoIce \cite{xiaoice}, and CASS \cite{cass}. Some of these models have gained massive success. For example, EMMA was incorporated in Xiaoai, the intelligent personal assistant created by Xiaomi Corporation and included by default in Xiaomi smartphones. As of 2023, Xiaomi is the third-largest seller of smartphones worldwide, with a market share of about 12\% \cite{counterpoint}. The chatbot XiaoIce, created by Microsoft Research Asia, enjoys similar success. As of 2020, it reached 660 million users and 450 million third-party smart devices globally \cite{xiaoice_users}. 


EMMA is an online empathetic chatbot which detects a set of 29 distinct emotion causes (e.g. breakup). It leverages two counseling strategies, effective questioning and active listening, to encourage self-disclosure and further engagement from the user. It then provides empathetic responses based on the detected emotion and its cause. Emotion recognition was treated as a multi-classification task; emotion cause was treated as a reading comprehension task, both performed using BERT \cite{bert}. User query, emotion, the presence of emotion cause, emotion cause and conversation history were concatenated and fed into a GPT model for training. 

XiaoIce is a seq2seq social chatbot that contains an empathetic computing module. It can recognize human feelings and states, understand user intent, and respond to user needs throughout long conversations. Its empathetic computing module classifies user intent into 11 existing categories such as greeting, user's emotion into five classes such as happy, and user's reaction to the current topic into positive, neutral and negative. These 3 pieces of information are concatenated with user query into a dialogue state vector which is used to select and activate a skill (e.g. core chat) through their hierarchical dialogue policy. The interpersonal responses are ultimately generated using a seq2seq model. 

CASS is a generalizable chatbot architecture to provide social support for members in an online health community for pregnant women. It first classifies posts into emotional-support and informational-support seeking categories; only emotional-support seeking posts are within its scope and kept. Then all emotional-support seeking posts and responses are fed into a seq2seq model with attention mechanism for training after advertisements and offensive posts are filtered out manually. 

Overall, these state-of-the-art empathetic open-domain chatbots leverage two key strategies for successful chatbot design: explicitness and large training datasets. Both EMMA and XiaoIce take an extra step to detect emotions and causes/intent and wire this information explicitly into the input before model training. CASS, however, does not have such explicitness and its success depends on whether the response posts from online forums are inherently empathetic, which is not guaranteed. 

With respect to the training dataset, XiaoIce claims to be the most popular social chatbot in the world and it has more than 30 billion conversation pairs as of May 2018. These volumes of data are far beyond what we (and many other researchers) have access to. CASS contains 220,000 post-response pairs. While the dataset of EMMA is smaller, containing 16,873 conversations, it employed a team of psychologists and human experts to write templates of active listening and effective questioning sentences and high-quality empathetic responses, thus making its dataset content-specific and empathy-guaranteed. By point of comparison, the dataset shared with us by PSI includes only 7,014 text conversations and 65,062 individual messages with a significant amount of ``logistics'' content when  volunteers are sharing local resources with support seekers (See Section~\ref{section: dataset}). This shrinks our useable dataset further, given that logistics content tends to be less empathetic and/or irrelevant to our chatbot design goals. Even with additional data scraping and collection, it would be hard to reach the volume of training data used by XiaoIce and CASS or the quality of training data used by EMMA for building a generative model. However, it is possible that supplementing a pre-trained domain-general generative model with a relatively small but domain-specific dataset could result in acceptable performance. Additionally, we leverage explicitness by developing natural-language processing models to detect the content of support seekers' concerns, thus allowing us to provide content-specific replies to support seekers. 

\section{Dataset} \label{section: dataset}
Through an extended collaboration with PSI, we were granted access to the content derived from the interactions between support seekers and trained volunteers on the PSI helpline. The PSI helpline serves as a 24/7 platform where emotional validation and resources are provided to postpartum women and families in need. Our access to this valuable dataset was made possible by signing a dataset transfer agreement, enabling us to acquire the complete text conversations from the helpline's database. For our study, we collected a total of 7,014 text conversations spanning the period from January 26, 2019, to October 12, 2020. These conversations contained 65,062 individual messages, with 29,108 (44.8\%) of them originating from support seekers seeking assistance. While geographical references were de-identified, it is worth noting that the helpline provided support across every state in the US, Canada, and internationally, wherever PSI Support Coordinators were stationed \cite{psi_2022}. Furthermore, an overwhelming majority of the helpline conversations, amounting to 99.6\%, were conducted in English.

We protect the privacy and anonymity of support seekers whose conversations we used to develop the chatbot in several ways. First, we de-identified all messages prior to analyses and modeling. Specifically, personally identifiable information within the conversations was identified using named-entity recognition techniques \cite{spacy} and subsequently replaced with standardized placeholders such as PSI\_PERSON and PSI\_PLACE. Next, all members of our research team certified that no efforts would be made to identify support seekers, and that the data would not be shared beyond our specifically named shared team. Additionally, in all published or publicly shared materials deriving from this dataset, all quotes from support seekers have been edited or paraphrased such that there is no direct relationship between the printed ``quotes'' and the material shared directly by support seekers, following e.g. \cite{diyi_onlineMedia}. Finally, our research was approved by the Institutional Review Board (IRB) at [redacted for blind review] and all researchers working on this project finished IRB training to ensure proper behaviors when handling sensitive text message content. 


\section{Descriptive Analysis of PSI Helpline Dataset} 
Previous work from Yao et al. \cite{descriptive_analysis} performed a descriptive analysis of the experiences of postpartum women using the PSI dataset we leverage to develop our chatbot. Our current work builds upon Yao et al.'s prior efforts and findings. Briefly, Yao et al. \cite{descriptive_analysis} found that there are stark differences between the concerns, psychological states, and goals of distressed and healthy mothers with distressed mothers describing interpersonal problems and a lack of support and the majority of healthy mothers describing childcare issues, such as questions about breastfeeding or sleeping. This informs us that distressed postpartum women need more targeted or concern-specific emotional validation from chatbots since their majors concerns are different from healthy parents. We leveraged the definitions and annotations of Yao et al. to build a detector to identify severe symptoms from support seekers. Additionally, we used Yao et al.'s analysis of the support seeker's concerns to identify the 13 most common concerns and 2 most common psychological states from distressed support seekers, and we developed classifiers corresponding to each of these concerns and psychological states. 

\begin{table}[!h]
  \caption{Clusters Generated from the Responses of PSI Volunteers with Definitions and Examples}
  \label{tab:clusters_psi}
  \begin{tabular}{p{5.5cm}p{7.5cm}}
    \toprule
    \textbf{Cluster Name} & \textbf{Examples} \\
    \midrule
    \textbf{Emotional Statements} & \\
    \hline
    Validation for a difficult time (9.64\%) & \textit{It sounds like you've got a lot on your plate not to mention everything going on in the world.} \\
    \hline
    Positive sentiment and validation (13.37\%) & \textit{Good for you for reaching out!} \\
    \hline
    Questions and validation (11.22\%) & \textit{Were you able to talk with your doctor today?} \& \textit{It can be really daunting sometimes, I hear you.} \\
    \hline
    PSI taglines  (9.23\%) & \textit{PSI is not a crisis line. } \& \textit{You are not alone, you are not to blame. With help you will be better. } \\
    \hline
    \textbf{Logistics} & \\
    \hline
    Introduction (6.25\%) & \textit{Hi this PSI\_PERSON, I'm a volunteer with PSI warmline.} \\
    \hline
    PSI info, limitations, and questions (5.19\%) & \textit{We are here to listen and connect you with resources, but we cannot diagnose or give advice regarding medications. } \& \textit{Do you have any support?} \\
    \hline
    Connecting to a local coordinator or therapist (16.08\%) & \textit{Ok I have reached out to our coordinator who specializes in this and I've asked her to be in touch as soon as possible.} \\
    \hline
    Assuring follow through (7.69\%) & \textit{I have reached out to your coordinator and you should hear from her within the next 24 hours.} \\
    \hline
    Online resources (5.24\%) & \textit{In the meantime, here are some additional resources through our website: [redacted]} \\
    \hline
    Crisis line info and local resources (16.09\%) & \textit{We aren't a crisis line, so if you are experiencing a crisis, please call PSI\_PHONE. They provide caring crisis support.} \\

    \bottomrule
\end{tabular}
\end{table} 

The focus of Yao et al.'s descriptive analyses was on the content shared by support seekers rather than volunteers. However, knowledge of the content shared by PSI helpline volunteers is also informative for chatbot development, given that this constitutes the training data for our generative chatbot's responses. Thus, we used the unsupervised clustering techniques described in Yao et al.'s work \cite{descriptive_analysis} to analyze the responses of PSI helpline volunteers. Clustering resulted in ten clusters as shown in Table~\ref{tab:clusters_psi} which can be roughly grouped into two categories, emotional statements (e.g. validation and encouragement) and logistics (e.g. providing resources and coordination). We find that in writing to support seekers, PSI volunteers responded more with logistics sentences than emotional statements (56.54\% vs 43.46\%). As machine learning highly depends on data, a training dataset skewed towards logistics will be better at providing logistics rather than empathy, which can be problematic. Thus, as detailed in Section~\ref{sec: gpt2chatbot}, we removed logistics inputs in an attempt to increase model performance.

\section{Models}
To develop and evaluate our chatbots, we trained 4 machine learning classifiers using the dataset provided by PSI. Our Severe Symptoms detection model, aims to identify severe symptoms from support seekers such that human volunteers instead of chatbots can provide them with precise and timely help.  Our Psychological States and Concerns detection models aim to identify the individual psychological states and concerns such that content-specific replies can be provided to support seekers. Finally, our Empathy detection model aims to identify and quantify the density of empathetic sentiment in a given response to evaluate the performance of chatbots developed.

\subsection{Severe Symptoms Detection Model} \label{sec: severe_symptoms_model}
Severe symptoms or behaviors include psychosis, suicidal thoughts, self-harm, and hurting others. A few examples include \textit{I wish I had never made it this long to watch myself fail.}, \textit{My break starts next week and I'm really considering ending it once and for all.} and \textit{I am afraid that I might hurt my baby and don't remember.} They are beyond the scope of our chatbots and thus we built a classifier to detect such occurrences.

To build a severe symptoms detection model, we first leveraged previous work \cite{descriptive_analysis} and collected a dataset containing annotated severe symptoms and other sentences generated by support seekers. Two types of features, sentence embeddings \cite{sbert} and LIWC \cite{liwc} features, were extracted and concatenated before being fed into a random forest classifier. For sentence embeddings, we leveraged the sentence-transformers model \footnote{\url{https://huggingface.co/sentence-transformers/all-MiniLM-L6-v2}} and mapped each annotated sentence to a 384 dimensional dense vector space. For LIWC features, we calculated whether each annotated sentence had words in 69 preset category of LIWC and generated a 69 dimensional binary vector. Data augmentation with back-translation \footnote{\url{https://huggingface.co/docs/transformers/model\_doc/marian}} was applied to balance the dataset. 3-fold cross validation was applied and the model performs well when identifying sentences containing severe symptoms with a mean F1 score 0.88 (Precision: 0.91, Recall: 0.85).

\subsection{Psychological States and Concerns Detection Models} \label{c: concern_model}
To provide content-specific replies to support seekers, we trained 2 classifiers, one to classify psychological states, and the other to classify the most common concerns, using the dataset annotated in previous work \cite{descriptive_analysis} and in the same way as our severe symptoms model. We determined the most common concerns based on the 13 most prevalent concerns among ``distressed'' mothers--support seekers in PSI dataset and r/ppd (see \cite{descriptive_analysis}). There are 4 psychological states in the annotated dataset. However, given that ``severe symptoms'' are out of the scope of our chatbot and ``unhealthy coping behaviors'' did not reach kappa (see \cite{descriptive_analysis}), we provide support to two psychological states in this chatbot: depressive mood and anxiety. The concerns and psychological states are listed in Table~\ref{tab:all_accuracy}.

\begin{figure}[!h]
\centering
\includegraphics[width=1.0\textwidth]{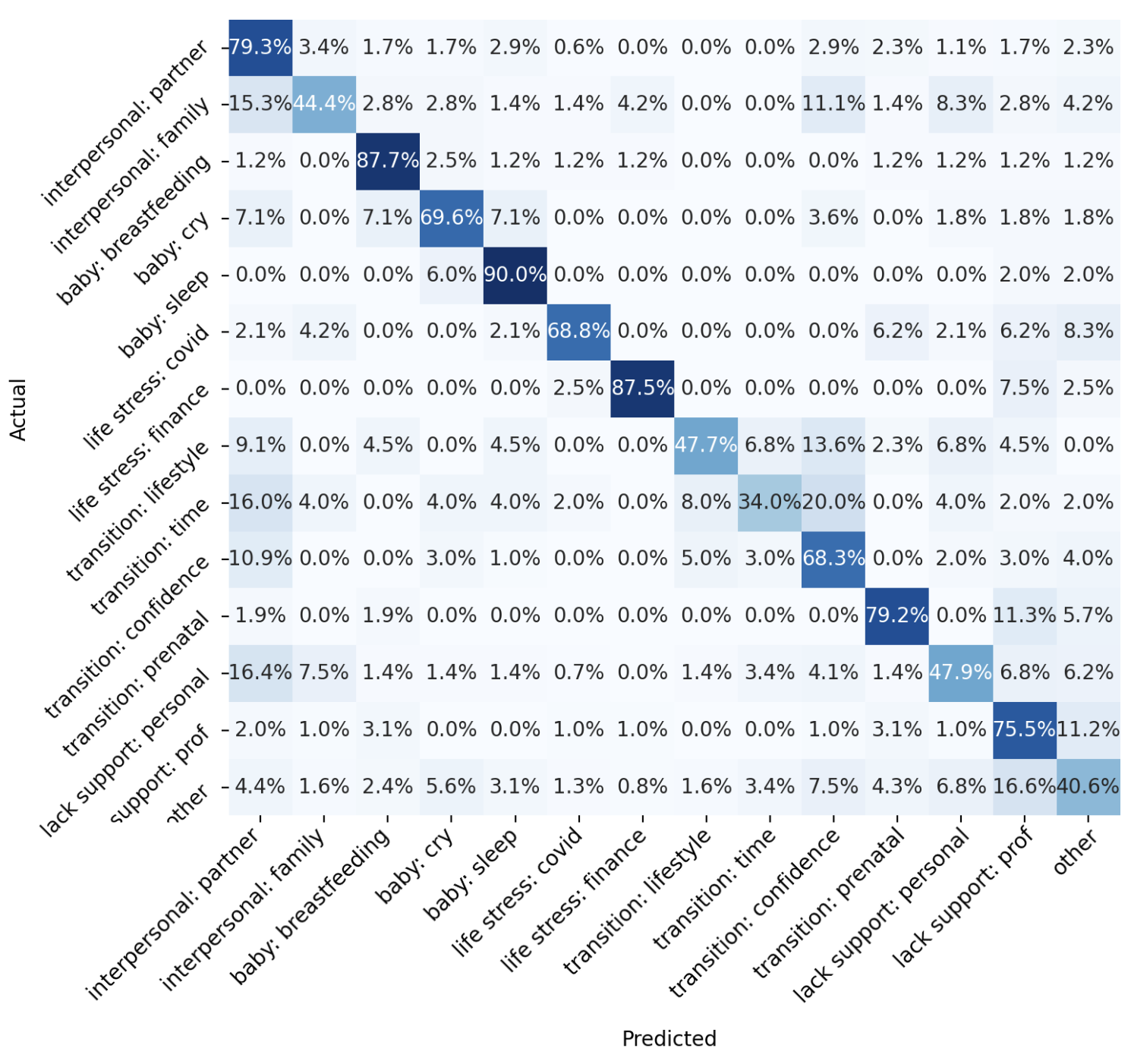}
\caption{Confusion Matrix for 13 Most Common Concerns Classification}
\label{fig:hardcoded_concerns_cm}
\end{figure}

\begin{figure}[!h]
\centering
\includegraphics[width=0.5\textwidth]{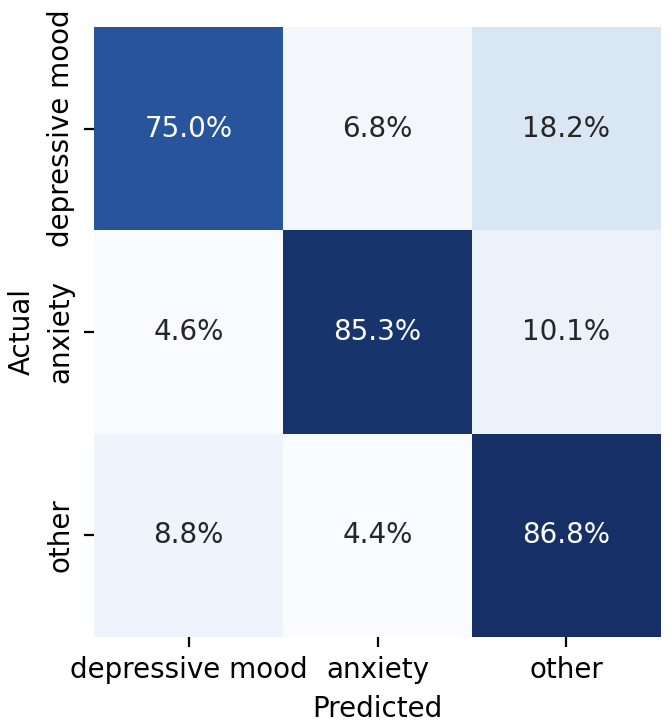}
\caption{Confusion Matrix for Psychological States Classification}
\label{fig:hardcoded_states_cm}
\end{figure}

The detailed accuracy and confusion matrices can be found in Table~\ref{tab:all_accuracy}, Figure~\ref{fig:hardcoded_concerns_cm}, and Figure~\ref{fig:hardcoded_states_cm}. We observe that our models have a high recall and a low precision for all concerns and psychological states, which is preferred as linguistic expressions are highly diverse. Therefore, chasing high precision can lead to fewer predictions and responses, resulting in less engagement and fewer possibilities to provide support. 

\begin{table}[ht]
\centering
  \caption{Classification Accuracy for 13 Most Common Concerns and 2 Psychological States}
  \label{tab:all_accuracy}
  \begin{tabular}{cccc}
    \toprule
     & \textbf{Precision} & \textbf{Recall} & \textbf{F1 Score }\\
    \midrule
    \rowcolor{Gray}
    interpersonal: partner  &  0.34  &  0.79  &  0.48 \\
    interpersonal: family  &  0.25  &  0.44  &  0.32 \\
    \rowcolor{Gray}
    baby: breastfeeding  &  0.36  &  0.88  &  0.51 \\
    baby: cry  &  0.13  &  0.70  &  0.21 \\
    \rowcolor{Gray}
    baby: sleep  &  0.22  &  0.9  &  0.35 \\
    life stress: covid  &  0.33  &  0.69  &  0.44 \\
    \rowcolor{Gray}
    life stress: finance  &  0.47  &  0.88  &  0.61 \\
    transition: lifestyle  &  0.20  &  0.48  &  0.28 \\
    \rowcolor{Gray}
    transition: time  &  0.09  &  0.34  &  0.15 \\
    transition: confidence  &  0.15  &  0.68  &  0.25 \\
    \rowcolor{Gray}
    transition: prenatal  &  0.17  &  0.79  &  0.28 \\
    lack support: personal  &  0.18  &  0.48  &  0.26 \\
    \rowcolor{Gray}
    lack support: prof  &  0.09  &  0.76  &  0.15 \\
    \hline 
    depressive moods & 0.58 & 0.75 & 0.66  \\
    \rowcolor{Gray}
    anxiety & 0.55 & 0.85 & 0.67 \\
    \bottomrule
\end{tabular}
\end{table}

\subsection{Empathy Detection Model} \label{c: empathy_model}
We created a metric Empathy\% to represent the percentage of output sentences containing empathy for evaluation. In order to calculate Empathy\%, we annotated empathy (emotional validation and encouragement) using the same dataset provided by PSI and built a classifier. The detailed annotation scheme and kappa \cite{kappa} for empathy are shown in Table~\ref{tab:annotation_empathy}. Both emotional validation and encouragement reach kappa higher than 0.8, indicating excellent agreement among raters \cite{why_0.4}. We also used the same classification method as our severe symptoms detection model to classify empathy against other sentences. Our empathy classifier reaches F1 score 0.86 (Precision: 0.82, Recall: 0.91), meaning that it can correctly detect empathy.

\begin{table}[!ht]
  \caption{Annotation Scheme for Empathy: Definition, Inter-rater Agreement (Kappa) and the Number of Conversations Containing the Type of Empathy}
  \label{tab:annotation_empathy}
  \begin{tabular}{p{3.3cm}p{7cm}lr}
    \toprule
     & \textbf{Definition \& Example} & \textbf{Kappa} & \textbf{\#}\\
    \midrule
    Validation & Reassurance of mother’s doubts or validate a fact/statement & 0.818 & 325\\
    Encouragement & Positive or hopeful statements about the future & 0.844 & 193\\
  \bottomrule
\end{tabular}
\end{table}

\section{Chatbots}
We describe the design and implementation of 3 chatbots with open-ended support seeker input, baseline (simple rule-based), rule-based (with NLP components), and generative, in this work leveraging the helpline dataset provided by PSI. As determined in collaboration with PSI, the design goals for all 3 chatbots are to provide i) empathetic and ii) content-specific responses that validate and address the specific concerns shared by support seekers, as well as iii) actively encourage support seekers to express and share their current challenges. Additionally, the chatbots should iv) refrain from offering advice or resources to support seekers, and v) refer support seekers suffering from severe symptoms to emergency services staffed by humans.

\subsection{Baseline Chatbot}
\begin{figure}[!h]
\centering
\includegraphics[width=0.8\textwidth]{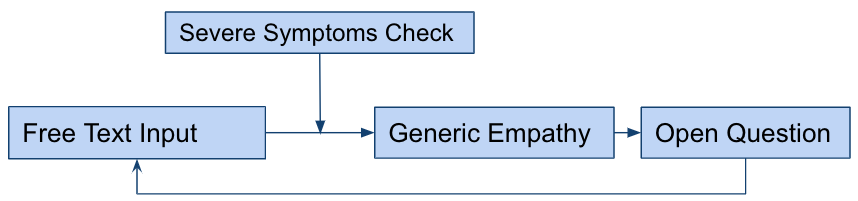}
\caption{The Flowchart of Baseline Chatbot}
\label{fig:baseline_flow}
\end{figure}

Our baseline chatbot uses a simple rule-based model that answers with generic, pre-written empathetic statements (emotional validation and/or encouragement) and asks open-ended questions to any text shared by a support seeker. Its flowchart is shown in Figure~\ref{fig:baseline_flow}. Table~\ref{tab:hardcoded_example} shows examples of generic empathy and open-ended questions and Table~\ref{tab: baseline_example_results} shows examples of input-output pairs of the baseline chatbot.

\begin{table}[h!]
\caption{Examples of Generic Empathy Sentences and Open Questions}
\label{tab:hardcoded_example}
\begin{tabular}{p{0.24\textwidth}p{0.7\textwidth}}
\hline
\textbf{Category} & \textbf{Example} \\
\hline
\multirow{6}{*}{Generic Empathy} & 	1. I would have a hard time with that too. \\
& 2. It’s got to be tough dealing with all of this. \\ 
& 3. Being a new parent is really hard and it sounds like you’re doing the best you can. \\ 
& 4. It sounds like you’re going through a lot. Try to be kind to yourself and know that you are not alone.\\
   
\hline
\multirow{4}{*}{Open questions} & 1. Do you want to talk more about that? \\
& 2. Tell me more about what happened? \\
& 3. Can you tell me a bit about what's going on? \\
& 4. What makes you feel this way? \\
\hline
\end{tabular}
\end{table}

\begin{table}[!ht]
  \caption{Example Outputs of Baseline Chatbot}
  \label{tab: baseline_example_results}
  \begin{tabular}{p{0.10\linewidth}|p{0.84\linewidth}}
    \toprule
    Input & I've been experiencing such intense anxiety that it feels like I'm dying.\\
    \hline
    \rowcolor{Gray}
     Baseline & That's all very understandable. 
     
     What makes you feel this way?
 \\
     \hline
     Input & I am very lonely throughout the day and I get upset with my husband. He's working from home but in another room. He didn't think about get dinner ready for me or the dog and instead he has been setting up his gym equipment in the garage. \\
     \hline
     \rowcolor{Gray}
     Baseline &  Being a new parent is really hard and it sounds like you're doing the best you can. 
     
     Can you tell me a bit about what's going on?
 \\
    \hline
    Input &  I have financial concerns all the time. I am worrying about not having enough college money for my baby or for him to get his first car. \\
    \hline
    \rowcolor{Gray}
    Baseline & That sounds really difficult. 
    
    Do you want to talk more about that?\\
  \bottomrule
\end{tabular}
\end{table}

\subsection{Rule-based chatbot}
Our rule-based chatbot aims to provide support seekers with pre-specified responses to their concerns and psychological states and asks support seekers open-ended questions. Ideally, we would be able to support postpartum women and provide related responses to all of their utterances. However, the concerns and psychological states represented in our data set are highly imbalanced such that there is not enough data to build a model that can detect each individual concern and psychological state (see \cite{descriptive_analysis}). Therefore, it was only feasible for us to build models that detect and provide high-quality responses to the most common concerns and psychological states. Thus, we developed two classifiers to detect the 13 most common concerns and 2 most common psychological states as detailed in Section~\ref{c: concern_model}.

\begin{figure}[!ht]
\centering
\includegraphics[width=1.0\textwidth]{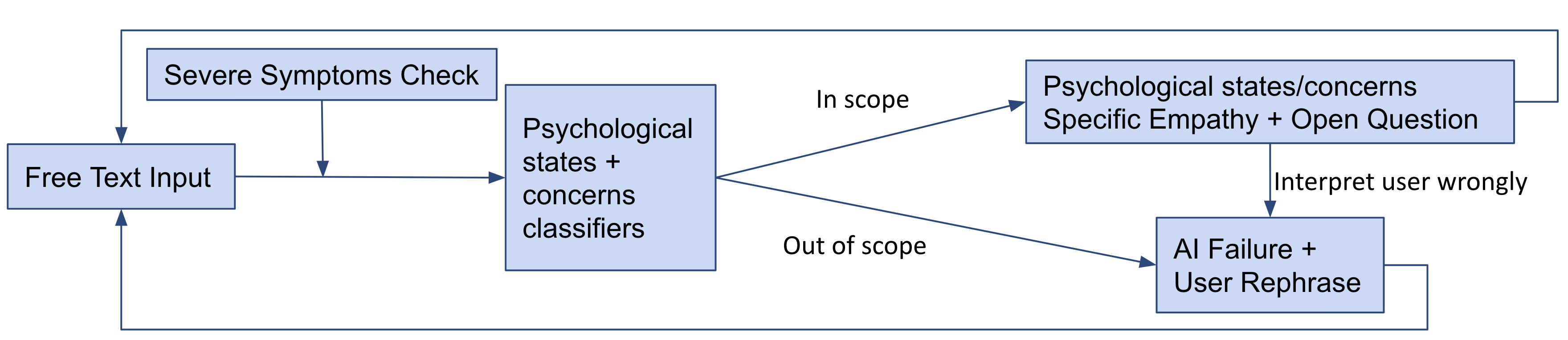}
\caption{The Flowchart of Rule-based Chatbot}
\label{fig:rulebased_flow}
\end{figure}

The flowchart of our rule-based chatbot is shown in Figure~\ref{fig:rulebased_flow}. Content-specific replies are randomly output from a pool of empathetic replies to the most common concerns and psychological states which we manually selected from the PSI dataset. Table~\ref{tab: rulebased_results} shows three examples of the outputs from rule-based chatbot. If the classifiers cannot detect concerns or psychological states, the user is notified of failure and can choose to rephrase their previous input. Otherwise, the chatbot outputs content-specific responses as well as open-ended questions. After receiving the reply, the user can signal if the chatbot incorrectly interpreted their input and can rephrase the previous input.


\begin{table}[!ht]
  \caption{Example Outputs of Rule-based Chatbot}
  \label{tab: rulebased_results}
  \begin{tabular}{p{0.10\linewidth}|p{0.84\linewidth}}
    \toprule
    Input & I've been experiencing such intense anxiety that it feels like I'm dying.\\
    \hline
    \rowcolor{Gray}
     Rule-based & It seems that you are feeling anxious.
     
    What you are feeling and going through is quite normal. Having a baby is a very hard transition for everyone....mind and body goes through a whole lot!
    
    Do you want to talk more about that?
 \\
     \hline
     Input & I am very lonely throughout the day and I get upset with my husband. He's working from home but in another room. He didn't think about getting dinner ready for me or the dog and instead he has been setting up his gym equipment in the garage. \\
     \hline
     \rowcolor{Gray}
     Rule-based &  It seems that you are having issues with your partner.
     
    I am sorry that you are not feeling connected to your partner. That's so frustrating. Parenthood is a partnership
    
    Tell me more about what happened?
 \\
    \hline
    Input &  I have financial concerns all the time. I am worrying about not having enough college money for my baby or for him to get his first car. \\
    \hline
    \rowcolor{Gray}
    Rule-based & It seems that you are feeling anxious and are having financial issues.

    Your financial situation sounds tough and I'm sure it's temporary and you will get the help you need.
    I know it feels like an injustice almost that the world just keeps spinning and everyone else seems to just be living their lives. It feels crushing and unfair and maybe even unbearable.

    Tell me more about what happened?\\
  \bottomrule
\end{tabular}
\end{table}

\subsection{Generative Chatbot} \label{sec: gpt2chatbot}
\begin{figure}[!h]
\centering
\includegraphics[width=0.7\textwidth]{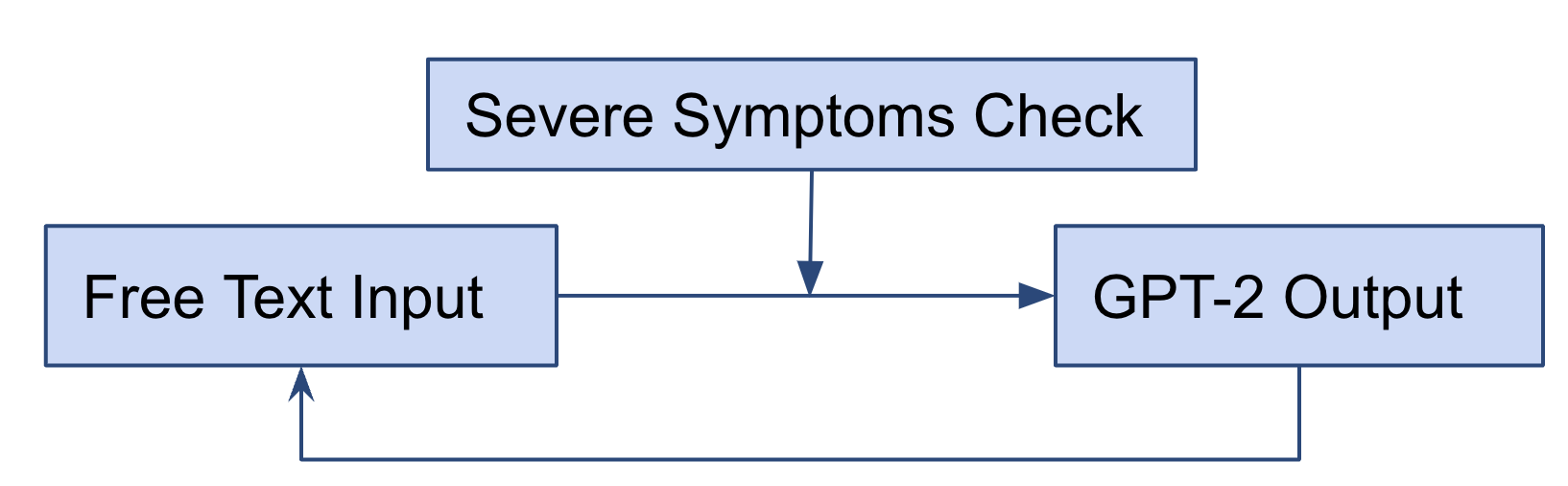}
\caption{The Flowchart of Generative Chatbot}
\label{fig:gpt_flow}
\end{figure}

To build our generative chatbot, we fine-tuned a GPT-2 Small language model \cite{gpt2} with the PSI dataset. We adopted an iterative development approach and built multiple versions of the chatbot throughout its implementation, detailed below. GPT-2 output is used directly as the chatbot output as shown in Figure~\ref{fig:gpt_flow}. We did not enforce the output of open-ended questions as we wanted the model to learn from the dataset.

\paragraph{Baseline Generative Chatbot} We built the baseline generative chatbot by fine-tuning unmodified GPT-2 Small \cite{gpt2} model with the PSI dataset, with the goal of establishing the baseline performance which we can improve upon with more complex future iterations. GPT-2 is a large pretrained transformer-based language model which generates the next word based on the current context. To ensure the quality of the training dataset, we excluded conversations that were short and had few conversational turns as those conversations typically do not provide useful interactions between support seekers and PSI volunteers. For example, in many short conversations, the support seeker initiates a conversation with a greeting but does not continue the exchange when the volunteer responds. Conversations with fewer than 3 conversational turns or 50 words total were removed from training, resulting in 4833 conversations and 14436 conversation turns, a reduction from 7014 conversations. 

We sampled 6 input-output pairs using top concerns and psychological states and found that in every case it could output coherent, relevant, and grammatically correct responses despite hallucination. However, 5 of the responses (83.3\%) appeared to be related to logistics, such as introduction and coordination with little empathy. This is expected since the majority of responses (56.54\%) in the training dataset are in the logistics cluster (see Section~\ref{section: dataset}). However, this feature does not meet our  stated design goal of providing empathetic responses to support-seekers' concerns. Additionally, given that the chatbot cannot actually provide logistical support, it is important that it does not ``suggest'' that it can in conversation with the support seekers. As such, in the next implementation of the generative chatbot, we retrained the model after removing logistics content.

\paragraph{Filtered Generative Chatbot}
In this implementation, we filtered the PSI dataset to exclude any sentence regarding logistic support and fine-tuned a filtered generative chatbot with this filtered dataset. To create the filtered dataset, we removed all PSI responses from the logistics clusters as well as all support seeker inputs from the coordination cluster, resulting in 8958 conversation turns (62.05\% of the original 14436 conversation turns in baseline chatbot). 

We first attempted to fine-tune a GPT-2 model by feeding the filtered dataset to an original pretrained GPT-2 model but the model did not output anything concerning postpartum women or empathy. We suspect that the small size of the filtered training dataset is not enough to fine-tune the GPT-2 model. To resolve this issue, we fine-tuned the baseline generative chatbot with the filtered dataset and created this filtered generative chatbot.

The example outputs of the filtered generative chatbot are shown in Table~\ref{tab: filtered_results}. We observe that every response contains empathy, as opposed to the baseline generative chatbot outputs which had empathy in 4 out of 6 responses (66.7\%). While the empathetic outputs from filtered generative chatbot are acceptable when the input is about general concerns or expressing negative moods, they do not target individual concerns, such as when a support seeker had financial concerns in the last example in Table~\ref{tab: filtered_results}. We hypothesize that this is because the concerns from support seekers are not evenly distributed, i.e. some concerns are less common than others, and the deficiency of certain concerns in the dataset leads to the model outputting generic responses. Additionally, the model outputs sentences with duplicated meanings, such as \textit{``I'm so glad you are reaching out. I'm so glad you are reaching out to us as well.''} This might be because the filtered dataset provides less variation in responses. Furthermore, the results still contain logistical responses as this model is fine-tuned on the baseline chatbot. Alternatively, this may be due to mistakes in the clustering model.

 \begin{table}[!ht]
  \caption{Example Outputs of Generative Chatbot}
  \label{tab: filtered_results}
  \begin{tabular}{p{0.1\linewidth}|p{0.84\linewidth}}
    \toprule
    Input & I've been experiencing such intense anxiety that it feels like I'm dying.\\
    \hline
    \rowcolor{Gray}
     Generative & Oh my goodness! What a difficult time! I can't believe you're having such a hard time! How old is your baby? \\
     \hline
     Input & I am very lonely throughout the day and I get upset with my husband. He's working form home but in another room. He didn't think about get dinner ready for me or the dog and instead he has been setting up his gym equipment in the garage. \\
     \hline
     \rowcolor{Gray}
     Generative &  I'm so sorry to hear that you're feeling this way. I'm sorry you have to suffer through this. I can help you find some resources on your own. \\
     \hline
     Input & I have financial concerns all the time. I am worrying about not having enough college money for my baby or for him to get his first car. \\
     \hline
     \rowcolor{Gray}
     Generative &   Hi, PSI\_PERSON. I'm so sorry you've been having a rough time. Would you like to share a little about what's going on?\\
  \bottomrule
\end{tabular}
\end{table}

\paragraph{Summary} Leveraging an iterative development strategy, we built 2 versions of a generative chatbot to provide support seekers with empathetic responses. Filtered generative chatbot is better as it contains more empathy (100\% vs. 66.7\%) and is used as the final version of generative chatbot. We notice that its empathetic responses are generic and have taken measures such as Plug and Play language model \cite{plug_and_play} to enforce the output of concern-specific responses. However, these attempts did not improve the performance of our model, possibly due to our small dataset size. Additionally, we note that the generative chatbot does not guarantee the output of open-ended questions, which is one of our goals for the chatbot.

\section{Evaluation and Results}
We conducted two sets of evaluations-- machine-based evaluation and human-based -- to compare the performance of three chatbots. We observe that the rule-based model achieves the best performance. Its outputs are close to ground truth reference and it contains the most empathy. Additionally, its outputs appeal more to human evaluators as they are more content-specific and human-like. Our generative chatbot provides empathetic responses, and users like that it is engaging and moves conversations towards problem resolution. However, it frequently outputs confusing responses, which limits its usage.

\subsection{Machine-based Evaluation}
\subsubsection{Comparison Datasets}
To evaluate our three chatbots systematically and automatically, we curated a dataset of PSI volunteer responses to create a comparison ``gold standard'' dataset. Working with PSI staff, we identified a list of PSI volunteers who were best at validating, empathetic conversations with support seekers. We randomly sampled conversational turns between those PSI volunteers and support seekers. As we designed our chatbots to provide empathy but not logistical support in non-emergency situations, when creating our gold standard reference dataset we removed conversational turns that i) did not contain empathy, ii) contained logistics, resources, advice, or iii) when support seekers displayed severe symptoms. The resulting gold standard dataset contained 100 input-output pairs representing best practices of empathetic responding from trained PSI volunteers. Additionally, we create an ``Average Standard'' dataset to evaluate the performance of chatbots against a ``typical'' or ``average'' support seeker in addtion to ``gold standard''. As a further comparison, we randomly sampled 100 input-output pairs from the original 7014 PSI conversations and similarly removed conversational turns that did not contain empathy or contained logistics, resources, advice, or when support seekers displayed severe symptoms. This dataset represents the average standard of the replies from PSI volunteers.

\subsubsection{Metrics}
We used two metrics, BERTScore \cite{bert-score} and Empathy\%, in the machine-based evaluation. As there are unlimited ways to express empathy in one's reply, we did not pick metrics focusing on syntactic similarity or exact match between our gold standard dataset and our chatbot responses, such as BLEU \cite{bleu} or ROUGE \cite{rouge}. Instead, we used BERTScore contextual embeddings to  calculate the semantic similarity between the gold standard dataset and our chatbot responses. We report three measures of similarity: Precision, Recall, and F1. Additionally, to capture and quantify empathy in chatbot replies, we built a classifier to detect the percentage of output sentences containing empathy (See Section~\ref{c: empathy_model}) and report the proportion of empathetic sentences using Empathy\%. As a point of comparison we also report Empathy\% for the gold standard dataset and the average standard datasets.

\subsubsection{Results}
Table~\ref{tab:machine_based_chatbot} shows the evaluation results using BERTScore (Precision, Recall, and F1) and Empathy\%. Overall, we observed that none of the three chatbots we develped were highly semantically matched to the gold standard PSI dataset, as evidenced by BERTScore F1 of less than 13\%. We hypothesize this is because of the diversity of language used to provide empathy. The rule-based model and the baseline model showed similar semantic matches that were both higher than our generative model. 
The generative model does not match semantically with the gold standard replies and has the lowest precision, recall, and F1 score. We speculate that this additional semantic mismatch is due to the large amount of ``logistic'' sentences in the output. Future work could remove ``logistic'' sentences from the training set after acquiring sufficient data. 


Empathy scores were overall much higher. First, we observed that the gold standard dataset had an Empathy\% of nearly 90\%, indicating that nearly 9 of 10 sentences in this dataset contained empathy. By contrast, the Average Standard dataset had an Empathy\% of 39.2\% with huge variance, implying stark inconsistencies in providing empathy in the replies of PSI volunteers. All three chatbots we built have Empathy\% between the Gold Standard and the Average Standard, with the rule-based model having the highest density of empathy (76.3\%), only  11.1\% lower than the gold standard, equivalent to one less sentence of empathy for every 10 sentences. The generative chatbot provides empathy at the lowest rate with only 53.6\% of the sentences containing empathy. However, given that all models have Empathy\% higher than the average standard we consider them acceptable. 


\begin{table*}
\centering
  \caption{BERTScore (Precision, Recall, F1) and Empathy\% (Percentage of Sentences Containing Empathy) for Baseline, Rule-based, and Generative Chatbot with Comparison Datasets}
  \label{tab:machine_based_chatbot}
  \begin{tabular}{ccccc}
    \toprule
     & \textbf{BERTScore} & \textbf{BERTScore} & \textbf{BERTScore} &   \multirow{2}{*}{\textbf{Empathy\%}} \\
     & \textbf{Precision} & \textbf{Recall} & \textbf{F1} & \\
    \midrule
    Baseline & 0.078($\pm$0.126) & 0.171($\pm$0.094) & 0.125($\pm$0.096) & 0.622($\pm$0.206)\\
    Rule-based & 0.177($\pm$0.102) & 0.076($\pm$0.088) & 0.127($\pm$0.075) & 0.763($\pm$0.164)\\
    Generative & 0.083($\pm$0.114) &  0.069($\pm$0.117) & 0.077($\pm$0.098) & 0.536($\pm$0.321)\\
    Gold Standard & - & - & - & 0.874($\pm$0.193) \\
    Average Standard & - & - & - & 0.392($\pm$0.312) \\

    \bottomrule
\end{tabular}
\end{table*}

\subsection{Human-based Evaluation}
After getting approved by IRB at [redacted for blind review], we recruited 14 PSI affiliates (volunteers and local coordinators) to evaluate and compare the performance of our three chatbots using a questionnaire. The questionnaire contains 14 5-point likert scale multiple choice questions aiming to measure the usability \cite{cuq}, usage \citep{urp_1, urp_2}, and satisfaction \cite{csq} of the chatbots as detailed in Table~\ref{tab:psi_evaluation} as well as four open-ended questions aiming to assess the advantages and disadvantages of each chatbot overall and relative to one another. PSI affiliates recruited had experience with the postpartum mood and anxiety disorders, all had children of their own, and often had initially connected to PSI via their own personal search for support. Additionally, through their work at PSI they had acquired a broader understanding of the concerns and psychological states faced by other postpartum women. Participants were instructed to interact with each chatbot in a given random order by prompting it as a typical postpartum support seeker looking for help during a 30-minute virtual session. After interacting with each chatbot, they were asked to complete the aforementioned questionnaire. The results are summarized in Table~\ref{tab:psi_evaluation} and Table~\ref{tab:psi_evaluation_text}.

Based on human evaluation, our rule-based chatbot performed best in every evaluation category and item. Participants preferred its detailed, human-like, and empathetic responses that were content-specific, i.e. matching the topics of user input (\textit{``The detailed conversation and more context around the questions made it seem that I am having a conversation with a real person''} and \textit{``Words of encouragement, sympathy and validation for the questions/concerns I share''}), which helps to \textit{``build rapport and make the user more trusting to share information while seeking help.''}

\begin{table*}
\caption{Chatbot Evaluation Survey (Multiple Choice Questions) and Results for Baseline, Rule-based, and Generative Chatbot}
\label{tab:psi_evaluation}
\begin{threeparttable}
\begin{tabular}{p{5.7cm}ccc}
\toprule
 & \textbf{Baseline} & \textbf{Rule-based} & \textbf{Generative} \\
\midrule
\textbf{Usability} & \textbf{3.22} & \textbf{4.00} & \textbf{3.19}\\
\hline
The interaction with PSIbot was realistic and engaging. & 3.57($\pm$1.12) & 4.21($\pm$0.86) &  3.71($\pm$1.03)\\
\hline
PSIbot understood me well. &  3.29($\pm$1.33) & 3.86($\pm$1.30)+ & 2.79($\pm$1.26)\\
\hline
PSIbot responses were useful, appropriate, and informative. &  3.50($\pm$1.24) & 4.00($\pm$1.31) &  3.29($\pm$1.33)\\
\hline
PSIbot appropriately identified where it could not understand me or help me. & 2.58($\pm$0.95)* & 4.15($\pm$0.86)+ & 3.00($\pm$1.08)\\
\hline
PSIbot is capable of holding a conversation. & 3.14($\pm$1.41) & 3.79($\pm$1.08) & 3.14($\pm$1.25))\\
\midrule
\textbf{Usage} & \textbf{4} & \textbf{4.34} & \textbf{4.07}\\
\hline
I understand how to use PSIbot without additional resources. & 3.71($\pm$1.22)& 4.14($\pm$0.99) & 3.92($\pm$1.07) \\
\hline
PSIbot is an affordable and accessible way to support postpartum women and families. & 3.93($\pm$1.10) & 4.36($\pm$0.89) & 4.07($\pm$0.88) \\
\hline
PSIbot encourages engagement. & 4.07($\pm$0.96) & 4.36($\pm$0.81) & 3.93($\pm$1.03)\\
\hline
The total time required to interact with PSIbot was reasonable. & 4.29($\pm$0.96) & 4.50($\pm$0.63) & 4.36($\pm$0.48) \\
\midrule
\textbf{Satisfaction} & \textbf{3.34} & \textbf{3.77} & \textbf{3.30}\\
\hline
How would you rate the quality of service you received?  & 3.43($\pm$1.05) & 3.93($\pm$0.70) & 3.50($\pm$0.98)\\
\hline
Do you feel PSIbot displayed adequate empathy? & 3.43($\pm$1.29) & 4.21($\pm$0.94)+ & 3.29($\pm$1.22) \\
\hline
To what extent has PSIbot met your needs? & 3.36($\pm$1.23) & 3.64($\pm$1.29) & 3.43($\pm$1.24) \\
\hline
If someone you know were in need of similar help, would you recommend PSIbot to them? & 3.14($\pm$1.30) & 3.57($\pm$1.18) & 3.14($\pm$1.30) \\
\hline
If you were to seek help again, would you come back to PSIbot? & 3.36($\pm$1.23) & 3.50($\pm$1.18) & 3.14($\pm$1.30) \\
\bottomrule
\end{tabular}
\begin{tablenotes}\footnotesize
    \item {\raggedleft \footnotesize   * denotes significant difference between Baseline and Rule-based \par}
    \item {\raggedleft \footnotesize   + denotes significant difference between Rule-based and GPT-2 \par}
    
   \end{tablenotes}
  \end{threeparttable}
\end{table*}

When it comes to disadvantages, PSI affiliates found the rule-based chatbot \textit{``frustrating due to the misinterpretations that occurred''}, i.e. when \textit{``responses were more complex but did not necessarily apply to my concerns''}. Another big issue according to PSI affiliates was its failure in providing practical advice and resources to help with concerns from support seekers, which is currently out-of-scope for our chatbots. Additionally, one participant found that it \textit{``at one point expand[ed] TOO much''}. This happened because the chatbot was programmed to provide concern-specific responses to each unique detected concern. Therefore, if six distinct concerns were detected in a support seeker's text, the chatbot would provide six concern-specific empathetic responses, one for each concern. Future work could target the key concerns (e.g. according to the number of times mentioned) and prioritize responses to shorten the reply. 

Overall, participants found our rule-based chatbot useful and helpful for people dealing with postpartum stress or depression as a new parent. It not only provides specific and empathetic responses consistently but also encourages users to explore their feelings and the reasons behind them.

Our human users ranked both the generative chatbot and the baseline chatbot consistently less strongly than the rule-based chatbots, with little difference in the score categories between these two chatbots. PSI affiliates confirmed that generative chatbot can provide specific and empathetic responses. The biggest advantage of generative chatbot, according to participants, was its ability to \textit{``move the conversation along''}. The training process of the generative chatbot involved ``logistics'' sentences from PSI, thus it had a self-introduction in the reply and moved the conversation from providing empathy towards resolution (giving advice or resources). This pipeline was \textit{``more approachable and less robotic''} according to PSI affiliates. However, the chatbot often did not provide any resources after first mentioning them, leaving participants confused. One possibility would be to train a detector to try to remove such content from generated responses. As we touch on in the Introduction, we are collaborating with PSI to provide support seekers with resources in a more robust way using hard-coded content in a distinct part of the proposed platform. Participants found the generative chatbot slighty less empathetic than the baseline chatbot, owing to the amount of ``logistics'' sentences provided. Most importantly however, open-ended responses suggested that the generative chatbot was \textit{``a bit confusing and out of place''} and \textit{``didn't seem to make sense in some of its responses''}. We suspect that the output of confusing responses is due to the small size of our training dataset. In summary, participants appreciated the natural flow of a conversation provided by the generative chatbot but recognized various flaws in the logical coherence of the responses themselves.

\begin{table}[!htb]
\caption{Chatbot Evaluation Survey (Text Entry Questions) and Results for Baseline, Rule-based, and Generative Chatbot}
\label{tab:psi_evaluation_text}
\begin{tabular}{cp{5.4cm}p{5.4cm}}
\toprule
 & \textbf{Advantages} & \textbf{Disadvantages} \\
\midrule
Baseline & 1. Quick, empathetic replies.

2. Open-ended questions that prompts in-depth answers. & 1. Robotic, repeated, bland, vague, generic, short replies.
 
2. Not provide resources or solutions.
 
3. Keep asking open-ended question after details. \\

\hline
Rule-based & 1. Empathetic, long, genuine, human-like, specific, encouraging responses.

2. Open-ended question. & 1. Not recognize some concerns. 
 
2. Not concise, sometimes too many responses.
 
3. No resources/advice.

4. Repeated response. \\
\hline
Generative & 1. Quick, empathetic, engaging, humanlike, less robotic response.

2. Move conversation along to resources, approachable. & 1. Confusing, not empathetic response. 

2. Mentioned resources but didn't provide any. \\
\bottomrule
\end{tabular}
\end{table}

PSI affiliates considered the baseline chatbot a good initial step to \textit{``figure out presenting issues for a help seeker''} and provide a \textit{``quick validation of concerns''}, however as the responses were short, robotic, generic, bland, and sometimes repetitive, PSI affiliates felt that it \textit{``didn't pick up on the conversation''} and \textit{``can only go so far with an individual before they would need a live person''}. Similar to the other two chatbots, PSI affiliates would like this chatbot to provide solutions to the concerns of support seekers instead of asking open-ended questions especially after details were already provided. They felt that \textit{``this would be problematic for new moms, especially since they have limited time and patience''}. Overall, PSI affiliates found the baseline chatbot less empathetic and its responses lacked depth, but it can be used as an initial step to gather information and get the stories of support seekers on paper.

\section{Discussions} \label{chatbot:discussion}
In this work, we designed, implemented, and evaluated three conversational agents, i.e. chatbots, to support caregivers experiencing or at risk of postpartum mood and anxiety disorders. We incorporated empathy and open-ended questions in our scheme and explored two common ways of building chatbots, rule-based models and generative models. Using both machine-based and human-based evaluations, we find that both rule-based and generative models are able to provide users with empathetic, engaging responses. However, our rule-based model scores higher in all three categories of human evaluation (usability, usage, and satisfaction) and is preferred relative to both the baseline and generative models for consistently providing human-like, content-specific empathetic responses. In contrast, the baseline model was perceived to be overly repetitive, and the generative model was inconsistent and often confusing in its responses. We conclude that a rule-based model is a better choice for supporting postpartum women experiencing or at risk of postpartum mood and anxiety disorders.


\subsection{Rule-based or Generative Model for Digital Mental Healthcare?}
In digital mental healthcare, rule-based models are currently more widely accepted and adopted compared to generative models, paralleling our results. Rule-based models not only are more frequently used in the wild \cite{survey_psychotherapy_chatbots} but also have passed randomized controlled trials in boosting people's mood and reducing people's symptoms \citep{woebot, wysa}, making their effects clinically relevant. 


The success of rule-based models partly derives from their use of pre-written responses, which ensures that the user will always receive an output that was developed and approved by researchers and/or clinicians. By contrast, generative models cannot guarantee that their responses would be approved by an expert in mental health care. The context in which a pre-written sentence is used can alter its meaning, potentially reducing its clinical effectiveness. However, overall this appears to be a ``safer'' option than relying on unverified outputs provided by a generative model, especially in the context of providing support to a vulnerable population. On the other hand, generative models have the potential to provide users with responses that are more customized and relevant to their inputs. Without the constraint of a pre-written pool of responses, a generative model can allow for conversations that have a more natural and human-like flow, which was a noted advantage in our human-based evaluation of generative chatbot. 

A big challenge with using generative models is their reliance on the dataset. Successful generative chatbots tend to have enormous training datasets as mentioned in Section~\ref{sec: c_relatedwork}. However, it is not always feasible to obtain such datasets in the context of psychotherapy due to privacy issues and ethical concerns. When a user's input is absent or not represented enough in the training dataset, the output from generative models can be unpredictable, confusing, and prone to errors such as hallucination, which is demonstrated in our generative chatbot. Another ethical concern for a generative model is if the training dataset is biased, it will be reflected in the chatbot responses. Our generative chatbot generated many ``logistic'' replies as they are represented in the training data, which is problematic given that it does not have the capability of verifying or connecting users to real-world resources. Thus, extra validation is required to prevent the model from outputting anything improper for the targeted sensitive population.

In summary, our work indicates that despite recent innovations in generative modeling, rule-based models are still the most suitable option for providing digital mental health support. In the future, generative models may be more successful given larger training datasets and sophiscated enough guardrails that guarantee the quality of the output from large language models.

\subsection{Ethical Concerns}
Given the sensitive nature of postpartum populations with potential mental health challenges, ethical issues will arise. First and foremost, it is crucial to provide the right support to individuals experiencing high levels of distress, including any indication of harming themselves or others. For this reason we developed a severe symptoms detection model. While our model had high accuracy, reaching F1 score 0.88, like all other machine learning models, it cannot assure complete accuracy. Thus, one recommendation would be to start or end each message with a disclaimer that the scope of the chabot is limited to issues that do not include severe symptoms, and providing resources for emergency services to all users. Another option would be to involve human intervention at least during the initial deployment with PSI.

Remaining ethical concerns are largely centered on the generative chatbot, given that the responses of the baseline and rule-based chatbot are selected from human responses. Since generative chatbot was trained using the PSI dataset, it often attempts to provide advice and resources, and it also occasionally asks users about their location or other personal information, counter to our design goals. Given that these are major goals of the PSI text message helpline (from which we collected our training data) this is not unexpected. In future efforts, detection models could be implemented to remove such responses. However, errors are inevitable, thus the current best approach is to have humans available to monitor the responses if generative models are to be implemented.


\subsection{Does ChatGPT Change the Game?}
Recently, large language models (LLMs), such as OpenAI's ChatGPT \cite{chatgpt} and Google's Bard \cite{bard}, have gained widespread attention. ChatGPT is the most prominent of these models and the fastest-growing consumer application in history, with 100 million monthly active users two months after launch \cite{chatgpt_users}. ChatGPT is fined-tuned from GPT-3.5 and uses Reinforcement Learning from Human Feedback (RLHF) with an astonishing 175 billion parameters and approximately 500 billion tokens in its training corpus \citep{chatgpt, instructGPT, gpt3}. In contrast, our generative chatbot was fine-tuned on GPT-2, an earlier and smaller version of GPT-3.5. Furthermore, the size of ChatGPT's dataset and its ability to generate coherent, grammatically correct, and meaningful sentences seem to address the key shortcomings we identified in our generative chatbot. However, regardless of the sophistication of ChatGPT's model, the question of whether it can provide consistently ethical and empathetic responses to vulnerable populations is less clear.

Undoubtedly, ChatGPT is an impressive demonstration of the state of the art in generative AI models. Its performance has spurred concern across industries about potential job replacement as well as discussions of its approach toward human intelligence \citep{chatgpt_exam, chatgpt_job}. The latter concern has elicited varying critiques which are particularly relevant to our question of ChatGPT's ability to generate human-like mental health support.


First and foremost, many experts argue that LLMs are further from reaching human intelligence than they may appear. In particular, Turing Award-winning Yann LeCun and postdoctoral researcher at New York University Jacob Browning \cite{ai_limits_language} stated that \textit{``a system trained on language alone will never approximate human intelligence''}. LLMs are proficient at finding high-order statistical patterns in large datasets and guessing the most likely next words and are effective predictive systems but those only give LLMs a shallow understanding as human beings share a non-linguistic understanding when communicating \citep{ai_limits_language, chatgpt_pinker}. As LLMs do not always have non-linguistic context explicitly in their training dataset, it will never ``understand''. Capacity for language does not imply thinking. Similarly, Mahowald et al. \cite{llm_cognitive} found that LLMs demonstrated impressive performance in formal linguistic competence, but failed in functional linguistic competence, such as formal reasoning, world knowledge, situation modeling, and social reasoning. Without an understanding of the situations or what empathy is, the reply simply provides the most probably next word and sentence which may not fit the context or contain empathy.

Additionally, we tested ChatGPT with typical prompts from postpartum women and provide extra reasons why it cannot be used in digital mental health yet.

\subsubsection{ChatGPT needs context to produce correct responses.} 
It has been reported that long chat sessions confused the model and caused it to become repetitive or prone to being ``provoked'' \cite{chatgpt_bing_news}, thus Bing reduced the limit of interaction to 5 conversation turns per session \cite{bing_reduce}. Our experiment shows that ChatGPT does not always remember previous conversations. With a user prompt \textit{``I am panicking that I hurt him and don't remember.''} in the middle of a long conversation, ChatGPT replied \textit{``The best way to do that is to reach out to the person directly and have an open and honest conversation with them.''} ChatGPT seems to ``forget'' that ``him'' in the context refers to an infant rather than an adult even though we chatted about it for the past several conversation turns. Such defect is not ideal when talking to ``distressed'' postpartum women or people with mental health challenges as they will not feel validated or understood.

\subsubsection{It is not ethical for ChatGPT to give advice.}
Prompting ChatGPT with baby crying concerns, it replies with a list of suggestions ranging form ``hold and rock your baby'' to ``seek support from a pediatrician''. Undeniably, those are terrific suggestions but it may not be proper for ChatGPT to provide advice. With the right prompt, ChatGPT can give correct specific steps for shoplifting and making explosives \cite{chatgpt_explosives}, serving as a reminder that ChatGPT is merely a LLM and it does not have human understanding or moral restraints. Additionally, information given by ChatGPT is not always factual and can be amazingly wrong. Stack Overflow has recently banned answers generated by ChatGPT, stating that the average rate of obtaining correct answers from ChatGPT is too low and undermines the trust within the community \cite{chatgpt_stackoverflow}. Thus without extra human verification, it is not ethical for ChatGPT to provide advice for digital mental healthcare.

\subsubsection{It is hard to update or fix a bug in ChatGPT}
It is not uncommon for chatbots to make mistakes or provide nonsensical replies, but it is hard for a generative language model to correct its mistake. Since all replies are pre-written for rule-based models, once a reply is deemed improper or obsolete, it can be removed or updated easily. LLMs have billions of parameters and developers do not know which parameters need tweaking for every mistake and retraining the model is computationally expensive \cite{llm_fix_update}. Researcher have developed measures, such as MEND \cite{llm_mend}, to update LLMs more easily, but as the real world is complicated and constantly progressing, these measures are not sustainable when a large number of updates are needed.  

To conclude, given that ChatGPT does not share a non-linguistic understanding with human beings, needs context to product correct responses, cannot provide advice ethically, and does not have a guarantee on how it corrects itself, LLMs like ChatGPT can be a helpful aid for therapists or peer supporters, but cannot be used in digital mental healthcare directly without human moderators.

\section{Conclusion}
In this work, we collaborated with Postpartum Support International and developed three chatbots to provide postpartum women with context-specific empathetic support. We explored both rule-based and generative models and evaluated the performance of our chatbots using both machine-based metrics and human evaluation. We found that the rule-based model achieved the best performance overall due to its content-specific and human-like replies. Generative models, on the other hand, output confusing or nonsensical responses due to the limitations in the training dataset. We conclude that large language models can be a helpful aid for therapists or peer supporters, but at this point they do not appear ready to be used in digital mental healthcare directly, that is, without human moderators.





\begin{acks}
This work was supported by the National Institute of Mental Health K01 Award (1K01MH111957-01A1) to Kaya de Barbaro and Good Systems, a research grand challenge at the University of Texas at Austin. We thank support seekers from PSI and Reddit for helping us get a better understanding of the experience of postpartum women and design a better version of one-to-one text platform.
\end{acks}

\bibliographystyle{ACM-Reference-Format}
\bibliography{sample-base}

\end{document}